\documentclass[conference,a4paper,english]{IEEEtran}[2015/08/26]
\usepackage{balance}
\usepackage{upquote}
\usepackage[ngerman,main=english]{babel}

\addto\extrasenglish{\languageshorthands{ngerman}\useshorthands{"}}
\usepackage[hyphens]{url}
\makeatletter
\g@addto@macro{\UrlBreaks}{\UrlOrds}
\makeatother
\usepackage[zerostyle=b,scaled=.75]{newtxtt}
\usepackage[T1]{fontenc}
\usepackage[
]{microtype}
\DisableLigatures{encoding = T1, family = tt* }
\usepackage{graphicx}
\usepackage{diagbox}
\usepackage{xcolor}
\usepackage{listings}
\usepackage{multirow}
\usepackage{multicol}
\definecolor{eclipseStrings}{RGB}{42,0.0,255}
\definecolor{eclipseKeywords}{RGB}{127,0,85}
\colorlet{numb}{magenta!60!black}
\lstdefinelanguage{json}{
    basicstyle=\normalfont\ttfamily,
    commentstyle=\color{eclipseStrings}, 
    stringstyle=\color{eclipseKeywords}, 
    numbers=left,
    numberstyle=\scriptsize,
    stepnumber=1,
    numbersep=8pt,
    showstringspaces=false,
    breaklines=true,
    frame=lines,
    string=[s]{"}{"},
    comment=[l]{:\ "},
    morecomment=[l]{:"},
    literate=
        *{0}{{{\color{numb}0}}}{1}
         {1}{{{\color{numb}1}}}{1}
         {2}{{{\color{numb}2}}}{1}
         {3}{{{\color{numb}3}}}{1}
         {4}{{{\color{numb}4}}}{1}
         {5}{{{\color{numb}5}}}{1}
         {6}{{{\color{numb}6}}}{1}
         {7}{{{\color{numb}7}}}{1}
         {8}{{{\color{numb}8}}}{1}
         {9}{{{\color{numb}9}}}{1}
}
\usepackage[autostyle=true]{csquotes}
\defineshorthand{"`}{\openautoquote}
\defineshorthand{"'}{\closeautoquote}
\usepackage{booktabs}
\usepackage{paralist}
\usepackage[%
  square,        
  comma,         
  numbers,       
  sort&compress  
]{natbib}

\usepackage{etoolbox}
\makeatletter
\patchcmd{\NAT@test}{\else \NAT@nm}{\else \NAT@hyper@{\NAT@nm}}{}{}
\makeatother
\usepackage{pdfcomment}

\usepackage{stfloats}
\fnbelowfloat
\usepackage[group-minimum-digits=4,per-mode=fraction]{siunitx}
\usepackage{hyperref}
\hypersetup{
  hidelinks,
  colorlinks=true,
  allcolors=black,
  pdfstartview=Fit,
  breaklinks=true
}
\usepackage[all]{hypcap}
\usepackage[caption=false,font=footnotesize]{subfig}
\usepackage[incolumn]{mindflow}
\usepackage[capitalise,nameinlink,noabbrev]{cleveref}
\crefname{listing}{Listing}{Listings}
\Crefname{listing}{Listing}{Listings}
\crefname{lstlisting}{Listing}{Listings}
\Crefname{lstlisting}{Listing}{Listings}
\usepackage{lipsum}
\usepackage[math]{blindtext}
\usepackage{mwe}
\usepackage[realmainfile]{currfile}
\usepackage{tcolorbox}
\tcbuselibrary{listings}
\DeclareFontFamily{U}{MnSymbolC}{}
\DeclareSymbolFont{MnSyC}{U}{MnSymbolC}{m}{n}
\DeclareFontShape{U}{MnSymbolC}{m}{n}{
  <-6>    MnSymbolC5
  <6-7>   MnSymbolC6
  <7-8>   MnSymbolC7
  <8-9>   MnSymbolC8
  <9-10>  MnSymbolC9
  <10-12> MnSymbolC10
  <12->   MnSymbolC12%
}{}
\DeclareMathSymbol{\powerset}{\mathord}{MnSyC}{180}

\usepackage{xspace}

\makeatletter
\newcommand{\hydash}{\penalty\@M-\hskip\z@skip}
\makeatother
\input glyphtounicode
\begin{document}
\title{Harnessing Machine Learning for Discerning AI-Generated Synthetic Images}

\author{%
  \IEEEauthorblockN{Yuyang Wang}
  \IEEEauthorblockA{Cornell University, USA\\
    \{yw2545\}@cornell.edu}
  \and
  \IEEEauthorblockN{Yizhi Hao}
  \IEEEauthorblockA{Cornell University, USA\\
    \{yz2222\}@cornell.edu}
  \and
  \IEEEauthorblockN{Amando Xu Cong}
  \IEEEauthorblockA{Cornell University, USA\\
    \{ax45\}@cornell.edu}
}
\IEEEspecialpapernotice{Final Project - CS 5785 Applied Machine Learning (2023FA)}
\maketitle

\begin{abstract}
In the realm of digital media, the advent of AI-generated synthetic images has introduced significant challenges in distinguishing between real and fabricated visual content. These images, often indistinguishable from authentic ones, pose a threat to the credibility of digital media, with potential implications for disinformation and fraud. Our research addresses this challenge by employing machine learning techniques to discern between AI-generated and genuine images. Central to our approach is the CIFAKE dataset, a comprehensive collection of images labeled as ``Real'' and ``Fake''. We refine and adapt advanced deep learning architectures like ResNet, VGGNet, and DenseNet, utilizing transfer learning to enhance their precision in identifying synthetic images. We also compare these with a baseline model comprising a vanilla Support Vector Machine (SVM) and a custom Convolutional Neural Network (CNN). The experimental results were significant, demonstrating that our optimized deep learning models outperform traditional methods, with DenseNet achieving an accuracy of 97.74\%. Our application study contributes by applying and optimizing these advanced models for synthetic image detection, conducting a comparative analysis using various metrics, and demonstrating their superior capability in identifying AI-generated images over traditional machine learning techniques. This research not only advances the field of digital media integrity but also sets a foundation for future explorations into the ethical and technical dimensions of AI-generated content in digital media.

\end{abstract}

\IEEEpeerreviewmaketitle

\section{Introduction and Motivation}
\label{sec:introduction}

\begin{figure}
    \centering
    \includegraphics[width=.8\linewidth]{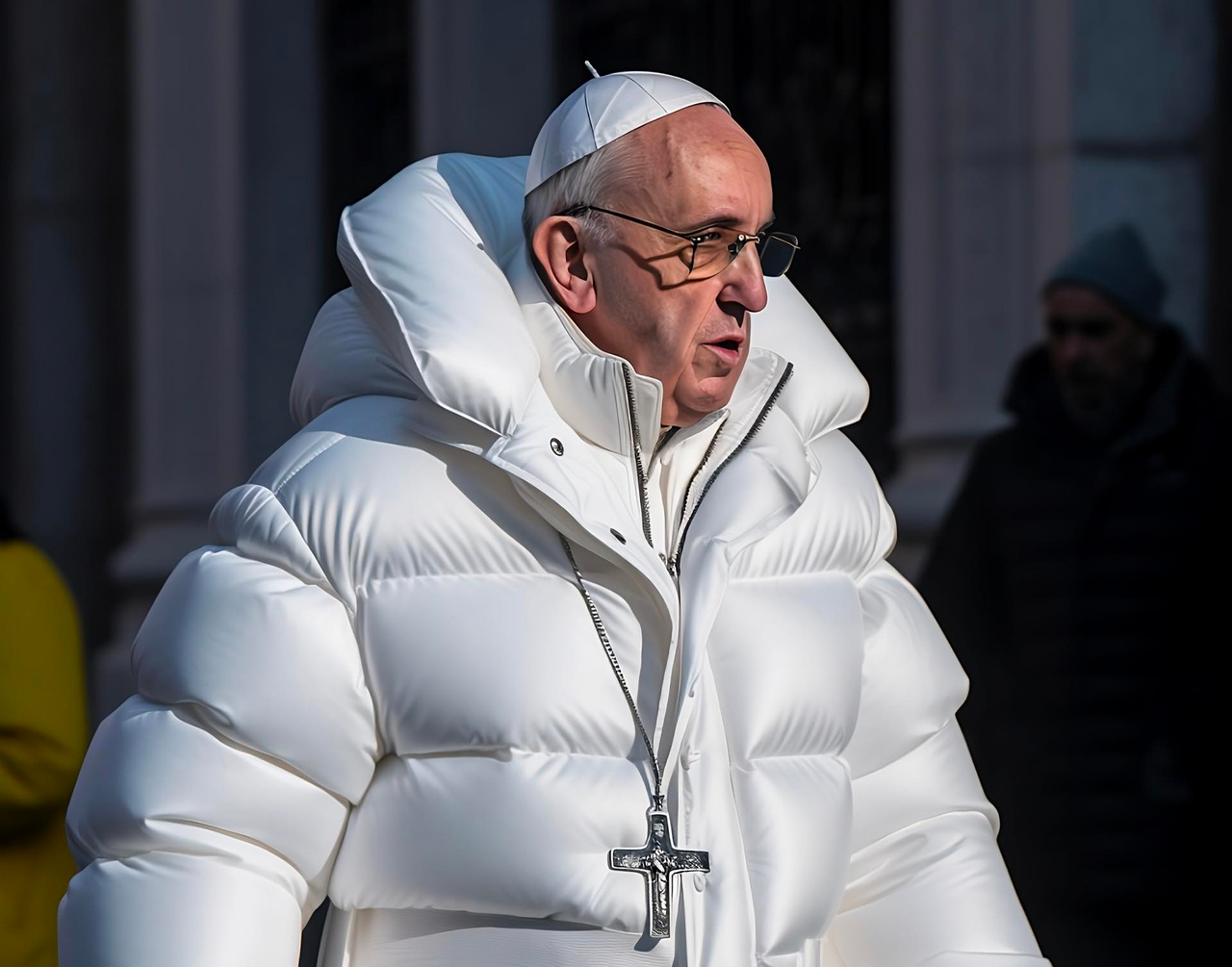}
    \caption{AI-generated Image of Pope Francis}
    \label{fig:pope}
\end{figure}

The proliferation of AI-generated synthetic images poses a significant challenge, blurring the lines between reality and digital fabrication. These images, such as the one shown in Figure \ref{fig:pope}, often indistinguishable from authentic ones, threaten the credibility of digital content and could potentially be exploited for disinformation and fraud. Addressing this issue is not just a technological challenge but a critical step towards preserving the integrity of digital media. Our study is an application project aiming to employ machine learning techniques to efficiently differentiate between AI-generated and genuine images. 

\begin{figure}
  \centering
  \includegraphics[width=.8\linewidth]{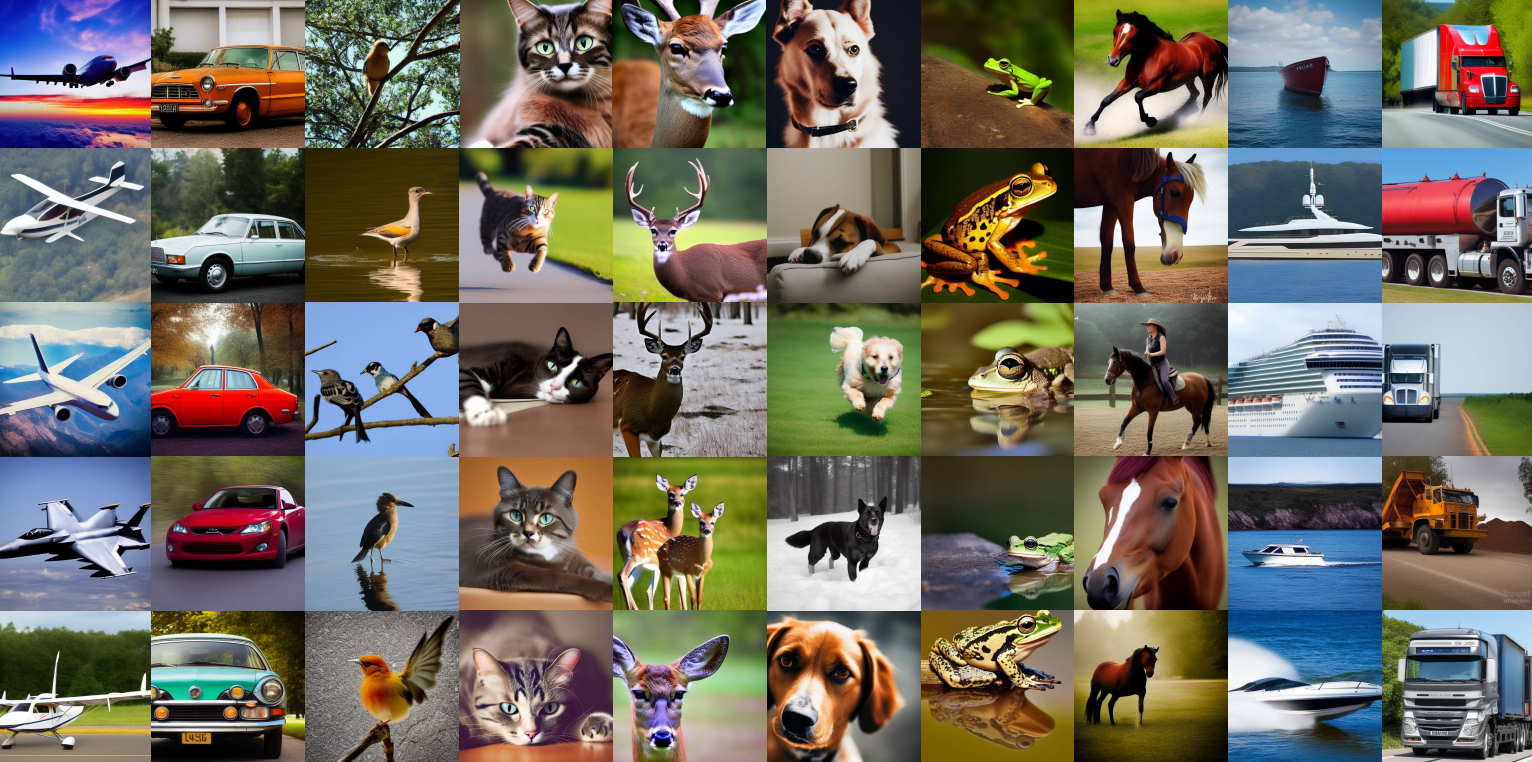}
  \caption{CIFAKE Dataset}
  \label{fig:CIFAKE Dataset}
\end{figure}

Central to our methodology is the CIFAKE dataset \cite{bird2023cifake}, exemplified in Figure \ref{fig:CIFAKE Dataset}, a comprehensive collection of images categorized as ``Real'' and ``Fake''. This dataset is used for training and testing our models. Our approach includes refining advanced deep learning architectures such as ResNet, VGGNet, and DenseNet through transfer learning. Each of these models, renowned for their effectiveness in image classification tasks, is adapted through transfer learning to suit our specific challenge. This process allows us to capitalize on the strengths of these pre-trained models, enhancing their precision and efficiency in identifying synthetic images. Additionally, we trained a vanilla method of Support Vector Machine (SVM) and designed a custom Convolutional Neural Network (CNN) to serve as baseline comparisons. These methods allow us to benchmark the efficiency and accuracy of our models in distinguishing synthetic images.

Our study is distinguished by several key contributions:
\begin{itemize}
  \item Application and optimization of advanced deep learning architectures (ResNet, VGGNet, DenseNet) for synthetic image detection.
  \item Development and training of a vanilla SVM and a custom CNN model as baseline methodologies.
  \item Conducting a comparative analysis of our models against existing methods using metrics such as accuracy, precision, recall, and F1-score.
  \item Demonstrating the enhanced capability of our optimized models in identifying AI-generated images, surpassing traditional machine learning techniques.
\end{itemize}

\section{Background}
\label{sec:background}




\subsection{Importance of Distinguishing Real from Fake}

The ability to discern real images from AI-generated ones is not just a technical challenge; it has broader implications for society\cite{1325569}. In an era where visual content plays a crucial role in shaping public opinion and personal beliefs, the reliability of digital information is paramount. Ensuring the authenticity of digital content is crucial in combating misinformation and maintaining trust in digital media.

\subsection{Evolution of Machine Learning in Image Analysis}

Machine learning, a subset of artificial intelligence, has seen rapid advancements in recent years. Its application in image analysis and recognition has been groundbreaking\cite{9391842}. Initially, traditional methods like Support Vector Machines (SVM) and basic neural networks paved the way for image classification tasks. However, the complexity and subtlety of AI-generated images necessitate more sophisticated approaches.

\subsection{Emergence of Deep Learning and CNNs}

Convolutional Neural Networks (CNNs) marked a significant leap in machine learning's ability to process and analyze visual data. Unlike traditional algorithms, CNNs can automatically and adaptively learn spatial hierarchies of features from image data\cite{9886935}. This capability is crucial in distinguishing subtle differences between real and artificially generated images.

\subsection{Advancements in Deep Learning}

Recent advancements in deep learning, particularly in architectures like Residual Networks (ResNet)\cite{he2015deep}, DenseNet\cite{huang2018densely}, VGGNet\cite{simonyan2015deep}, and Inception, offer new possibilities. These models, known for their deep structures and complex feature extraction capabilities, provide a more refined approach to discerning between real and synthetic images. The application of these models, coupled with techniques like transfer learning, presents a promising frontier in the fight against digital disinformation.

\subsection{CIFAKE Dataset}

CIFAKE dataset, a comprehensive collection of images specifically designed for training and testing AI models in distinguishing between real and fake images. This dataset's diversity and volume make it an ideal benchmark for evaluating the effectiveness of various machine learning models.

\section{Method}
\label{sec:mthod}


This section outlines the methodological approach undertaken in this study to discern AI-generated synthetic images from authentic ones. The machine learning algorithms employed, their configuration, and the rationale behind their use are detailed.

\subsection{Support Vector Machine (SVM)}
The Support Vector Machine (SVM)\cite{708428} was selected as the initial classifier to establish a comparative baseline for the neural network models. Recognized for its proficiency in categorizing high-dimensional data into distinct classes, SVM is apt for binary classification of images. Classification is achieved through hyperplanes in a multidimensional space, optimized by solving the following primal problem:
\begin{equation}
\min_{\mathbf{w}, b} \frac{1}{2} ||\mathbf{w}||^2 + C \sum_{i=1}^n \xi_i
\end{equation}
subject to \( y_i (\mathbf{w} \cdot \mathbf{x}_i + b) \geq 1 - \xi_i \), where \( \xi_i \) are slack variables allowing misclassification. The Radial Basis Function (RBF) kernel \( K(\mathbf{x}_i, \mathbf{x}_j) = \exp(-\gamma ||\mathbf{x}_i - \mathbf{x}_j||^2) \) addresses non-linear complexities in image data. A soft margin approach allows a degree of misclassification, enhancing the SVM's performance and robustness against diverse image scenarios.

\subsection{Convolutional Neural Network (CNN)}
Convolutional Neural Networks (CNNs)\cite{oshea2015introduction} form the core of our image classification investigation. Structured to automatically detect and learn features from raw image pixels, CNNs utilize convolutional layers with filters that capture spatial hierarchies, defining each layer operation as:
\begin{equation}
y_{m,n} = f_{m,n} * k_{m,n} = \sum_i \sum_j f_{i,j} \cdot k_{m-i, n-j}
\end{equation}
where \( f \) is the input image and \( k \) is the kernel function. The ReLU activation function \( \text{ReLU}(z) = \max(0, z) \) and pooling operations such as \( \text{MaxPool}(A) = \max(A) \) reduce dimensionality and introduce non-linearity. Dropout and Batch Normalization layers are incorporated to prevent overfitting and accelerate convergence.

\begin{figure}
    \centering
    \includegraphics[width=.9\linewidth]{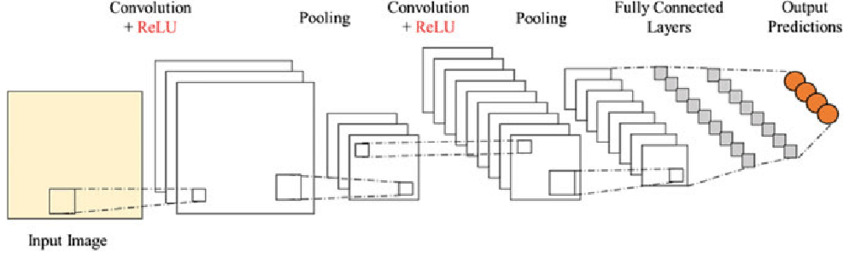}
    \caption{Typical CNN Model}
    \label{fig:Typical CNN Model}
\end{figure}

\subsection{Residual Network (ResNet)}
The Residual Network (ResNet) architecture is employed to train deep networks, countering the vanishing gradient problem with 'residual blocks' that include skip connections:
\begin{equation}
\mathbf{x}_{l+1} = \mathbf{x}_l + \mathcal{F}(\mathbf{x}_l, \mathbf{W}_l)
\end{equation}
where \( \mathbf{x}_l \) is the input to the \( l^{th} \) layer, \( \mathcal{F} \) is the residual function, and \( \mathbf{W}_l \) are the weights. This enables the training of networks with hundreds of layers, crucial for complex classification tasks.

\subsection{Visual Geometry Group Network (VGGNet)}
The Visual Geometry Group Network (VGGNet) is adopted for its deep architecture, characterized by multiple convolutional layers with small receptive fields, enabling systematic feature capture. This depth allows VGGNet to process detailed nuances within the image datasets, learning nuanced representations essential for accurate classification.

\subsection{Densely Connected Convolutional Networks (DenseNet)}
The Densely Connected Convolutional Networks (DenseNet) augment the depth of VGGNet with an architecture that connects each layer to every subsequent one, enhancing feature reuse and reducing parameter count:
\begin{equation}
\mathbf{x}_l = \mathcal{H}([\mathbf{x}_0, \mathbf{x}_1, \dots, \mathbf{x}_{l-1}])
\end{equation}
where \( \mathcal{H} \) represents composite functions of batch normalization, ReLU, pooling, and convolution, and the concatenation \( [\mathbf{x}_0, \mathbf{x}_1, \dots, \mathbf{x}_{l-1}] \) denotes the accumulated feature maps. DenseNet's efficient connectivity and depth offer a potent approach for distinguishing subtle differences in the dataset of real and synthetic images.

\section{Experimental analysis}
\label{sec:experimentAlanalysis}

\subsection{Dataset}

Our study primarily utilizes the CIFAKE dataset \cite{bird2023cifake}, which is a substantial collection of AI-generated synthetic and authentic images. This dataset consists of 120,000 images categorized into ``Fake'' and ``Real''. The images in CIFAKE were carefully curated to ensure high-quality and reliable labeling, making it an ideal choice for our machine learning endeavor. The dataset was split as follows for our experimental setup:

\begin{table}[h]
    \centering
    \caption{CIFAKE Dataset Splits}
    \begin{tabular}{lcc}
      \toprule
      & \textbf{Training} & \textbf{Testing} \\
      \midrule
      \textbf{``Fake'' Images} & 50,000 & 10,000  \\
      \textbf{``Real'' Images} & 50,000 & 10,000 \\
      \textbf{Total Images} & 100,000 & 20,000 \\
      \bottomrule
    \end{tabular}
    \label{tab:dataset_splits}
\end{table}

\subsection{Hardware and Software}

Our experimental work was mainly executed on Google Colab, which provided a conducive environment for team collaboration and also offered GPU resources, significantly speeding up our model training. Regarding software tools, we utilized Python for the entire range of our experimental tasks, encompassing data preprocessing, model training, and performance evaluation. Specifically, we implement the SVM model using \textit{Scikit-Learn}. For the development of CNN, ResNet, VGGNet, DenseNet, we opted for \textit{PyTorch}.

\begin{table*}[h]
    \centering
    \caption{Classification Reports for Different Models}
    \begin{tabular}{l|ccc|ccc|ccc}
      \toprule
      \multirow{2}{*}{\textbf{Model}} & \multicolumn{3}{c|}{\textbf{Fake}} & \multicolumn{3}{c|}{\textbf{Real}} & \multicolumn{2}{c}{\textbf{Overall}} \\
      & \textbf{Precision} & \textbf{Recall} & \textbf{F1-Score} & \textbf{Precision} & \textbf{Recall} & \textbf{F1-Score} & \textbf{Accuracy} & \textbf{ROC-AUC}\\
      \midrule
      SVM          & 0.82 & 0.80 & 0.81 & 0.81 & 0.83 & 0.82 & 0.81 & 0.90\\
      CNN          & 0.86 & 0.87 & 0.87 & 0.87 & 0.85 & 0.86 & 0.86 & 0.93\\
      ResNet       & \textbf{0.99}$\uparrow$ & 0.91 & 0.95 & 0.91 & \textbf{0.99}$\uparrow$ & 0.95 & 0.95 & \textbf{0.99}$\uparrow$\\
      VGGNet       & 0.97 & 0.95 & 0.96 & 0.95 & 0.97 & 0.96 & 0.96 & \textbf{0.99}$\uparrow$\\
      DenseNet     & 0.98 & \textbf{0.98}$\uparrow$ & \textbf{0.98}$\uparrow$ & \textbf{0.98}$\uparrow$ & 0.98 & \textbf{0.98}$\uparrow$ & \textbf{0.98}$\uparrow$ & \textbf{0.99}$\uparrow$\\
      \bottomrule
    \end{tabular}
    \label{tab:classification_reports}
\end{table*}

\subsection{SVM}

\subsubsection{SVM Data Processing}

Images are loaded and converted to grayscale to simplify the feature extraction process. The images are then resized to the uniform size of 32x32 pixels, ensuring consistency in input dimensions. The Histogram of Oriented Gradients (HOG) method is employed to extract features from these images. The features are further normalized using a Standard Scaler, ensuring that they follow a standard Gaussian distribution, which is beneficial for the convergence of the SVM algorithm.

\subsubsection{SVM Model}

We employ the Radial Basis Function (RBF) kernel for the SVM model, which is suitable for handling non-linear relationships in the data. The model is trained and subsequently evaluated on the test dataset to obtain various performance metrics. 

\subsubsection{SVM Results and Analysis}

The SVM model achieved an accuracy of 81.43\% on the test dataset, demonstrating a reasonable level of effectiveness for the task. The confusion matrix, as shown in Table \ref{tab:svm_confusion_matrix}, indicates a balanced classification capability between the two classes. The classification report, as shown in Table \ref{tab:classification_reports} shows a slightly higher precision for classifying 'Fake' images and a marginally better recall for 'Real' images. Both classes have similar F1-scores, indicating a balanced trade-off between precision and recall. Overall, the model exhibits a balanced performance with an accuracy of 81.43\%. The ROC-AUC score of 0.8958 and the Precision-Recall AUC of 0.8907 are also indicative of the model's relatively good discriminative ability and reliable classification performance in an imbalanced dataset scenario.

\begin{table}
    \centering
    \caption{Confusion Matrix for SVM Model}
    \begin{tabular}{lll}
      \toprule
      & \textbf{Predicted: Fake} & \textbf{Predicted: Real} \\
      \midrule
      \textbf{Actual: Fake} & 8020 & 1980 \\
      \textbf{Actual: Real} & 1734 & 8266 \\
      \bottomrule
    \end{tabular}
    \label{tab:svm_confusion_matrix}
\end{table}

\subsection{CNN}


\subsubsection{CNN Data Processing}
Similar to SVM, the images are converted to grayscale and resized a dimension of 32x32 pixels. Unlike the SVM approach, raw pixel values are utilized as features instead of applying the HOG method, as the convolutional layers in the CNN are capable of feature extraction.

\subsubsection{CNN Model Architecture}

\begin{figure}
    \centering
    \includegraphics[width=0.9\linewidth]{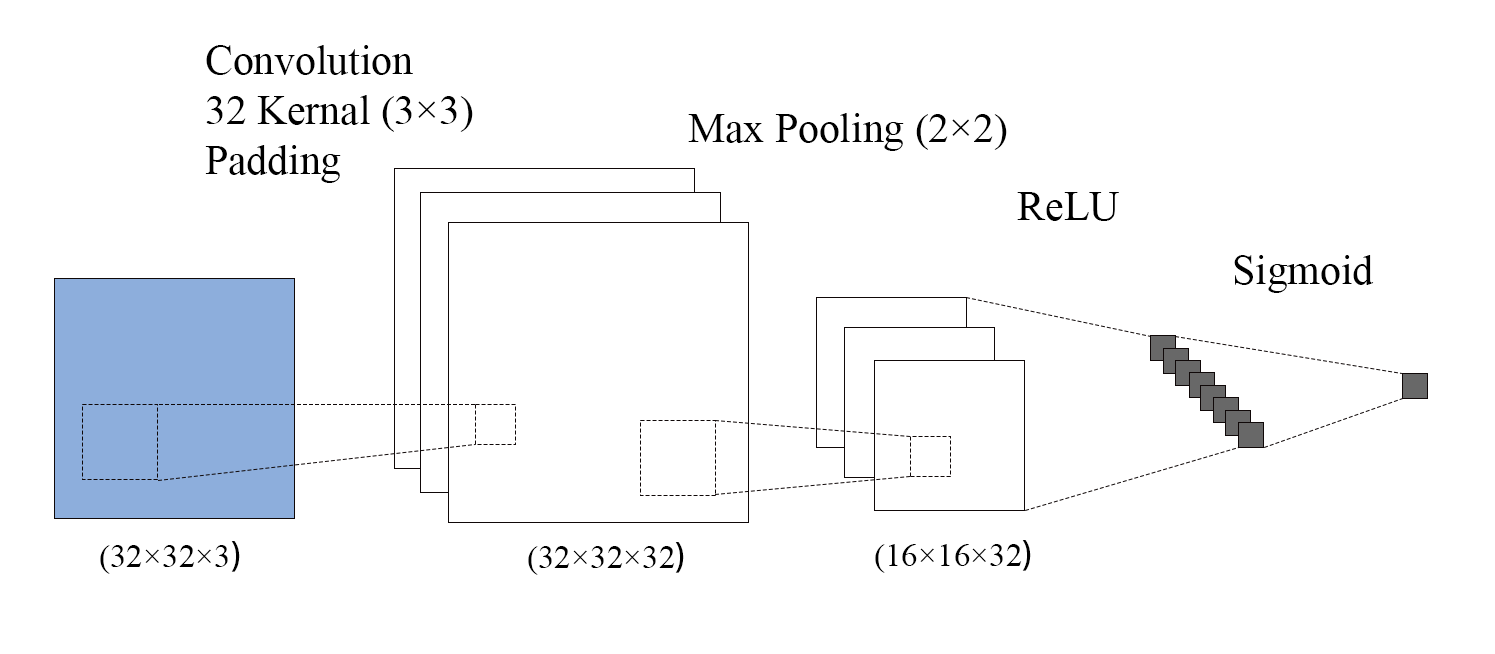}
    \caption{The Architecture of the Custom CNN Model}
    \label{fig:customcnn}
\end{figure}

As shown in Figure \ref{fig:customcnn}, the architecture of our custom CNN model includes an initial convolutional layer with 32 filters of kernel size 3x3, followed by max pooling of kernel size 2x2 with 2 strides and no padding, and lastly two fully connected layers. The model employs ReLU activation functions and a sigmoid function in the fully connected for binary classification.

\subsubsection{CNN Model Training}

The model was trained using the Adam optimizer with a learning rate of 0.01. Training was conducted over 25 epochs with a batch size of 64. During training, the model's parameters were updated to minimize the cross-entropy loss. The training process was logged, and the model's accuracy and loss were monitored to track its learning progress.

\subsubsection{CNN Results and Analysis}

\begin{figure}
    \centering
    \includegraphics[width=.7\linewidth]{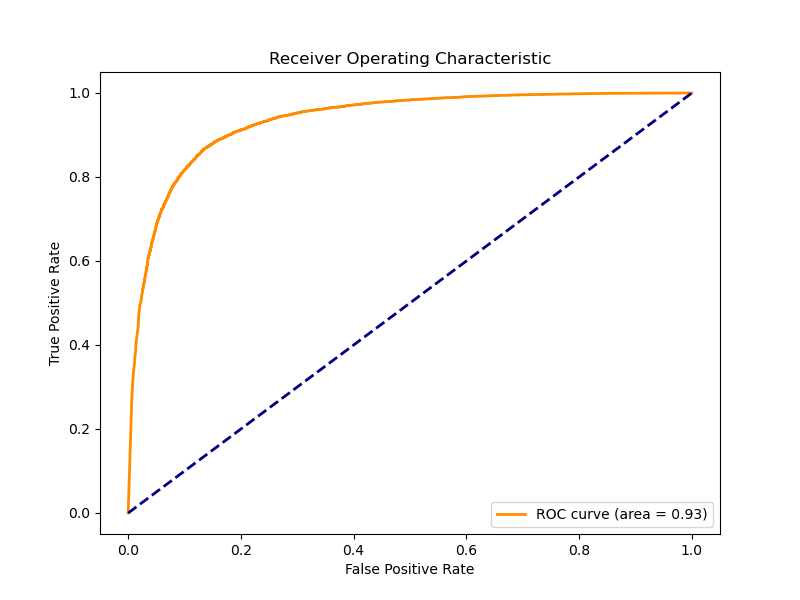}
    \caption{CNN Model ROC Curve}
    \label{fig:cnn_roc_curve}
\end{figure}

The CNN model showed significant improvement over the epochs, reaching an accuracy of 86.40\% on the test dataset. The confusion matrix, as shown in Table \ref{tab:cnn_confusion_matrix}, demonstrates the model's balanced classification capability for both 'Fake' and 'Real' images. The classification report, as shown in Table \ref{tab:classification_reports} further affirms the model's effective performance. The ROC-AUC score of 0.9347 indicates the model's strong discriminative ability and reliability in classifying images in the CIFAKE dataset. The ROC curve of the CNN model is shown in Figure \ref{fig:cnn_roc_curve}.

\begin{table}[h]
    \centering
    \caption{Confusion Matrix for CNN Model}
    \begin{tabular}{lcc}
      \toprule
      & \textbf{Predicted: Fake} & \textbf{Predicted: Real} \\
      \midrule
      \textbf{Actual: Fake} & 8734 & 1266 \\
      \textbf{Actual: Real} & 1453 & 8547 \\
      \bottomrule
    \end{tabular}
    \label{tab:cnn_confusion_matrix}
\end{table}

\subsection{ResNet}

\subsubsection{Data Processing}

\begin{itemize}
    \item Training Data Transformations
    \begin{itemize}
        \item {Random Resized Crop}: Each image in the training set is randomly cropped to a size of 24x24 pixels. This randomness in cropping ensures that the model is exposed to different parts of the images, enhancing its ability to generalize from the training data.
        \item {Random Horizontal Flip}: Images are randomly flipped horizontally, adding variance to the dataset and further aiding in generalization.
        \item {Normalization}: The images are normalized using mean values of [0.485, 0.456, 0.406] and standard deviations of [0.229, 0.224, 0.225]. This step is crucial for ResNet, which has been pre-trained on ImageNet where this normalization was applied.
    \end{itemize}
\end{itemize}

\begin{itemize}
    \item Training Data Transformations
    \begin{itemize}
        \item {Resize and Center Crop}: For the testing set, each image is resized to 32 pixels along the shortest side and then cropped from the center to 24x24 pixels. This ensures consistency in image size while evaluating the model.
        \item {Normalization}: The same normalization as applied to the training set is also applied here to maintain consistency in data preprocessing.
    \end{itemize}
\end{itemize}

\subsubsection{ResNet Model}

\begin{figure}
  \centering
  \includegraphics[width=1\linewidth]{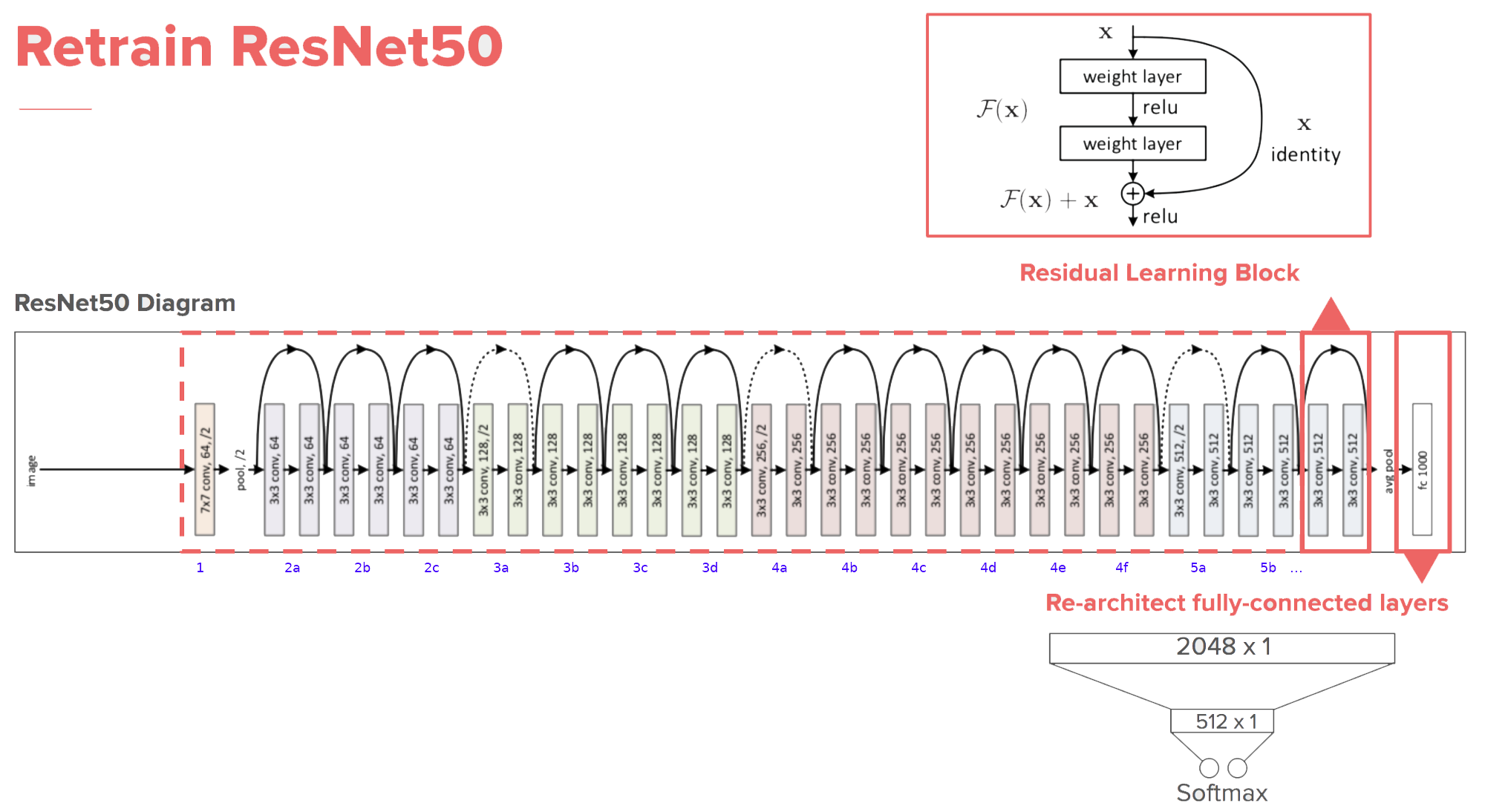}
  \caption{Modified ResNet Model}
  \label{fig:Modified ResNet Model}
\end{figure}

We begin by initializing the ResNet-50 model with weights pre-trained on the ImageNet dataset. This is achieved using \texttt{models.resnet50(pretrained=True)}. Leveraging a pre-trained model provides a significant advantage as it has already learned a wide range of image features, which can be fine-tuned for our specific task.

The model's fully connected (fc) layer, originally designed for ImageNet's 1000 classes, needs to be modified for our binary classification task (real vs. fake). We first extract the number of input features to the fc layer using \texttt{model\_ft.fc.in\_features}. This gives us the size of the tensor input to the fc layer.

Adapting the Output Layer: The next step involves replacing the existing fc layer with a new one tailored to our needs. Since we have a binary classification task, the output size is set to 2 (indicating two classes: real and fake). This is implemented via \texttt{nn.Linear(num\_ftrs, 2)}, where \texttt{num\_ftrs} represents the number of input features to the layer. It's noteworthy that while we specify '2' for our binary classification, this can be generalized to any number of classes by replacing '2' with \texttt{len(class\_names)}.

\subsubsection{ResNet Hyper-parameters}

\begin{itemize}
    \item {Loss Function}: We use the Cross-Entropy Loss, implemented as \texttt{nn.CrossEntropyLoss()}. This loss function is particularly suited for classification tasks, as it measures the performance of a model whose output is a probability value between 0 and 1. Cross-Entropy Loss increases as the predicted probability diverges from the actual label, making it an effective measure for our binary classification problem.
    \item {Optimizer}: we opted for the Adam optimizer, implemented as \texttt{optim.Adam(model\_ft.parameters(), lr=0.001)}. Adam, an algorithm for first-order gradient-based optimization, is chosen for its adaptive learning rate capabilities and efficient computation. A learning rate of 0.001 is selected to ensure a gradual and steady convergence during training.
    \item {Learning Rate Scheduler}: The learning rate scheduler, \texttt{lr\_scheduler.StepLR(optimizer\_ft, step\_size=10, gamma=0.1)}, is employed to adjust the learning rate during training. This scheduler reduces the learning rate by a factor of 0.1 every 10 epochs. Such a decay in the learning rate helps the model to converge more effectively by taking smaller steps when approaching a minimum in the loss landscape.
\end{itemize}

\subsubsection{ResNet Results and Analysis}

The ResNet model demonstrated a marked proficiency in classification tasks, achieving an accuracy of 94.95\% on the test dataset, which significantly surpasses that of the SVM and our custom CNN, highlighting the advanced capabilities of deep learning architectures and transfer learning in complex classification tasks. The classification report highlights its robustness, with precision of 91\% for real images and 99\% for fake images, and recall rates of 99\% for real and 91\% for fake images, leading to a balanced F1-score of 95\% for both classes. Furthermore, the model's ROC-AUC score of 0.9958 emphasizes its exceptional discriminative ability. These superior performance metrics, particularly in precision, recall, and ROC-AUC scores, clearly demonstrate the ResNet model's advanced capabilities in image classification, surpassing simpler models such as SVM and the self-designed CNN, and marking it as a more effective tool for this specific task.

\begin{table}[h]
    \centering
    \caption{Confusion Matrix for ResNet Model}
    \begin{tabular}{lcc}
      \toprule
      & \textbf{Predicted: Fake} & \textbf{Predicted: Real} \\
      \midrule
      \textbf{Actual: Fake} & 9,066 & 76 \\
      \textbf{Actual: Real} & 934 & 9,924 \\
      \bottomrule
    \end{tabular}
    \label{tab:resnet_confusion_matrix}
\end{table}

\begin{figure}
  \centering
  \includegraphics[width=.8\linewidth]{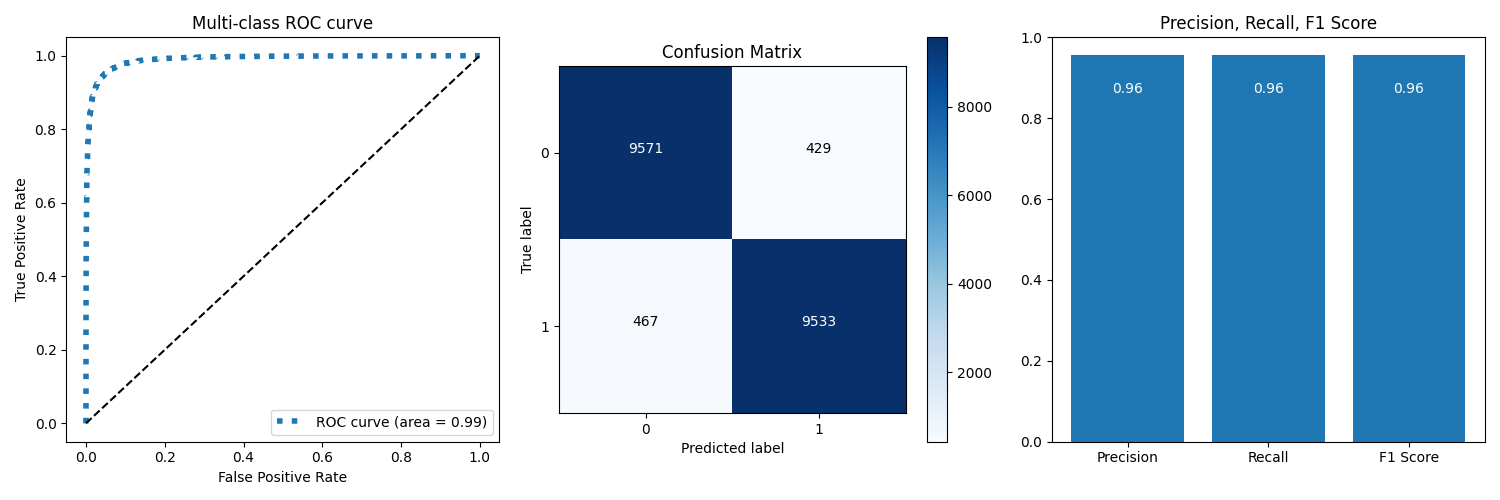}
  \caption{ResNet ROC curve}
  \label{fig:ResNet ROC curve}
\end{figure}

\subsection{VGGNet}

The VGGNet model, following the same data preprocessing as the ResNet, exhibits impressive performance on our task, which slightly outperforms ResNet. Throughout the training process, the model's accuracy consistently improved, starting at 84.23\% in the initial epoch and reaching approximately 91.72\% by the $20^{th}$ epoch. In testing, the model's accuracy started at 93.28\% and increased to about 96.00\% in the final epoch, with the highest validation accuracy recorded at 95.99\%. This indicates the model's effectiveness in accurately classifying images. 

\begin{figure}
  \centering
  \includegraphics[width=.7\linewidth]{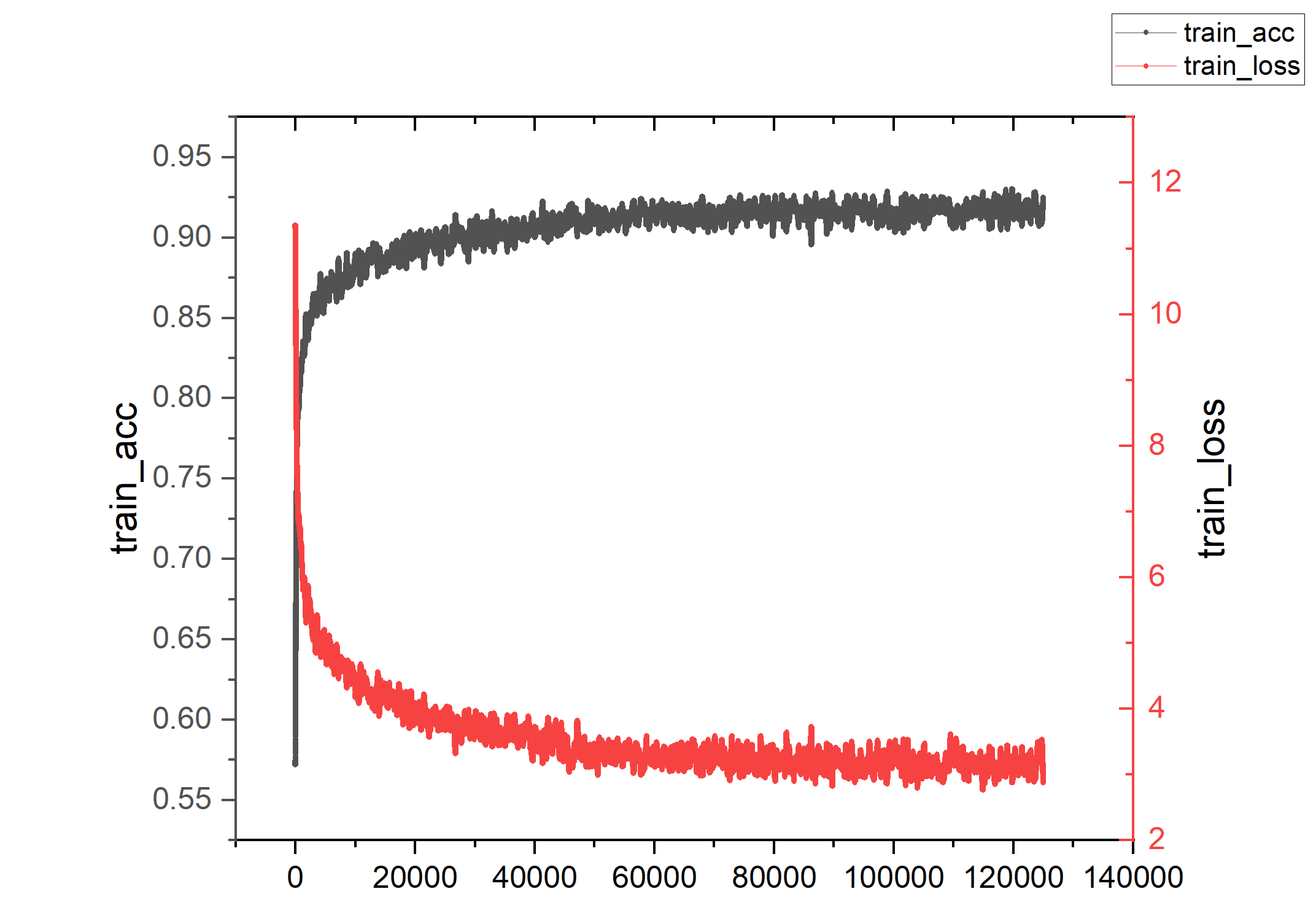}
  \caption{VGGNet Train Training Accuracy and Loss}
  \label{fig:VGGNet Train Result}
\end{figure}

In terms of class-specific performance, VGGNet identified 9,542 true positives and 458 false negatives for class 0, and 9,657 true positives and 343 false positives for class 1. This resulted in a final accuracy of 95.99\% on a test set of 20,000 images, underscoring its capability to differentiate between various image categories effectively. The VGGNet model's robust performance, marked by a steady increase in accuracy and a decrease in loss over time, as shown by Figure \ref{fig:VGGNet Train Result}, indicates effective learning and adaptation by the model to the image dataset.

\subsection{DenseNet}

\begin{figure}
    \centering
    \includegraphics[width=.7\linewidth]{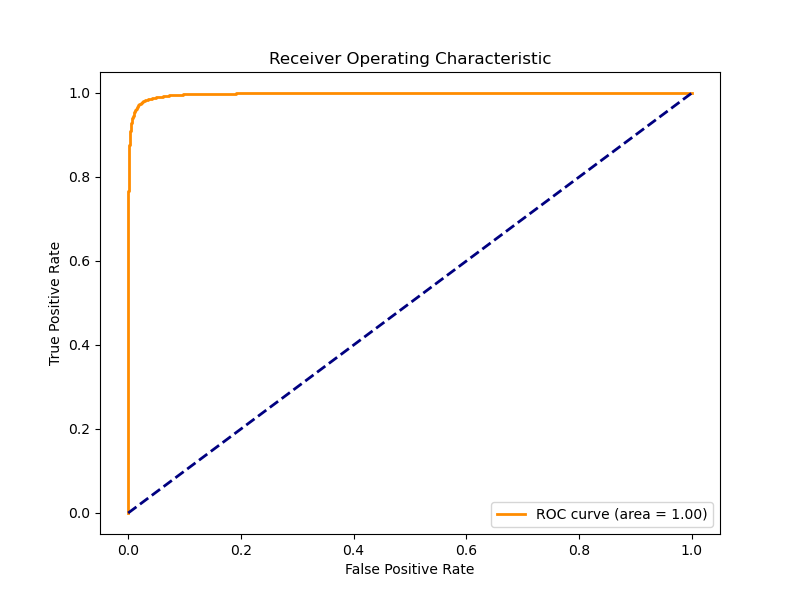}
    \caption{DenseNet Model ROC Curve}
    \label{fig:densenet_roc_curve}
\end{figure}

The pre-processing and implementation of the DenseNet follows a similar pipeline to the ResNet model. In terms of its performance, the DenseNet model demonstrated superior performance compared to other models as shown in Table \ref{tab:classification_reports}. The model achieves an accuracy of 97.74\% on the test dataset. The confusion matrix, as shown in Table \ref{tab:densenet_confusion_matrix}, along with the ROC-AUC score of 0.9975, and Precision-Recall AUC of 0.9976 reflect the model's great discriminative ability and reliability in classifying images in the CIFAKE dataset. The ROC curve of the DenseNet model is illustrated in Figure \ref{fig:densenet_roc_curve}.

\begin{table}[h]
    \centering
    \caption{Confusion Matrix for DenseNet Model}
    \begin{tabular}{lcc}
      \toprule
      & \textbf{Predicted: Fake} & \textbf{Predicted: Real} \\
      \midrule
      \textbf{Actual: Fake} & 9769 & 231 \\
      \textbf{Actual: Real} & 221 & 9779 \\
      \bottomrule
    \end{tabular}
    \label{tab:densenet_confusion_matrix}
\end{table}

\section{Discussion and Prior Work}
\label{sec:discussionAndPriorWork}

\subsection{Reflection on Prior Work}
Our study, focusing on the CIFAKE dataset\cite{bird2023cifake}, builds upon the foundation laid by previous research in the field of AI-generated image detection. It contributes to the field by comparing established machine learning methods on the CIFAKE dataset\cite{bartosdeep}. This comparison is crucial as it benchmarks the performance of traditional models like SVM and more advanced deep learning models such as ResNet and VGGNet against a unique dataset\cite{bhinge4594547quantifying}. Our findings resonate with existing research that underscores the effectiveness of deep learning in complex image recognition tasks but provide new insights by applying these models to a different kind of challenge involving synthetic media\cite{vora2023classification}.

\subsection{Key Takeaways}
A significant takeaway from our study is the diverse performance capabilities of various models when applied to the task of distinguishing real from synthetic images. Our analysis illustrates the remarkable capability of deep learning models, especially DenseNet, in accurately identifying AI-generated images. This aligns with the current trajectory in AI research, where deep learning continues to set benchmarks in image recognition tasks\cite{hinterstoisser2018pre}. Additionally, our study revealed the importance of dataset selection and preprocessing techniques in model performance, underscoring the need for carefully curated datasets like CIFAKE. While our study presents an initial exploration, it contributes to the broader understanding of how different machine learning models perform in a controlled academic setting\cite{meagher1982efficient}.

\section{Conclusion}
\label{sec:conclusion}

In this study, we explored the application of various machine learning models, including SVM, CNN, ResNet, VGGNet, and DenseNet, in distinguishing between real and AI-generated synthetic images. Our experiments demonstrated that while traditional methods like SVM provide a solid baseline, advanced models like DenseNet offer superior accuracy, underscoring the efficacy of deep learning in more complex image classification scenarios.

Future research could delve deeper into optimizing these models for tasks involving synthetic media, exploring the integration of new deep learning architectures, and extending the dataset to include more varied examples of AI-generated content. Additionally, understanding the ethical implications and potential misuse of AI in creating synthetic media is an important direction. Our study, while focused on a specific dataset and task, opens the door for more comprehensive research in the field of AI and digital image authenticity.

\bibliographystyle{IEEEtranN} 
\bibliography{paper}

\ \\
%

\end{document}